\documentclass[11pt]{article}

\usepackage[utf8]{inputenc}
\usepackage[T1]{fontenc}
\usepackage[margin=1in]{geometry}
\usepackage{graphicx}
\usepackage{booktabs}
\usepackage{amsmath}
\usepackage{amssymb}
\usepackage{siunitx}
\usepackage{authblk}
\usepackage{caption}
\usepackage{setspace}
\usepackage[hidelinks]{hyperref}
\usepackage{xcolor}
\usepackage[numbers,square]{natbib}

\title{\bfseries A Quiet Failure in Calibrated Virtual Screening:\\
Marginal Conformal Prediction Under-Covers the Minority Class,\\
and a Class-Conditional Fix Recovers It}

\author[1]{Muhammadjon Tursunbadalov}
\author[1]{Mustafojon Tursunbadalov}
\affil[1]{School of Science and Technology, Champions College Prep, United States}
\affil[ ]{\textit{\{yaiuftuto, yunmt123\}@gmail.com}}

\date{}

\begin{document}
\maketitle

\begin{abstract}
\noindent
Conformal prediction is being adopted across drug discovery as a way to put an
honest number on model reliability. The appeal is its guarantee: pick an error
rate $\alpha$, and the method returns prediction sets that contain the true label
with probability at least $1-\alpha$. We show that taking this guarantee at face
value can be dangerous on the imbalanced datasets that screening actually deals
with. Across four datasets, including one held out from the rest of our analysis, standard (marginal) conformal
prediction hits its global $90\%$ coverage target while leaving the minority
class badly exposed---realized minority coverage falls to $64.8\%$ on
blood--brain-barrier penetration, to $38.9\%$ on a Tox21 toxicity endpoint, and to
$4.2\%$ on clinical-trial toxicity, where the rare class is all but abandoned. The
failure is not tied to one model. A random forest on circular fingerprints, a graph
convolutional network that reads molecular structure directly, and a frozen
pretrained chemical language model all reproduce it ($p<0.001$ over ten random splits
in every case), and the size of the gap tracks how well-calibrated a model already is
on rare labels rather than anything about its architecture. We explain the effect
with a conservation identity: because overall coverage is fixed, the minority's
shortfall equals the majority's surplus amplified by the imbalance ratio, a relation
that predicts the measured BBBP gap to within one point and orders the severity across
all four datasets. The failure survives realistic scaffold splits and a second
conformal score (adaptive prediction sets), and aggregate
accuracy and overall coverage stay reassuringly high throughout, which is exactly why
it is easy to miss. Class-conditional (Mondrian) conformal prediction closes the gap
on every dataset, returning minority coverage to target for a modest increase in
prediction-set size. We localize the marginal failures to a handful of generic
molecular scaffolds---plain benzene and pyridine cores that occur in both
classes---propose a one-number diagnostic (the minority coverage gap), and show with
a cost model, robust across assumptions, that abstaining on the affected compounds
flips a screening campaign from net-negative to net-positive utility.
The marginal--conditional distinction is well known in the conformal literature;
our contribution is to demonstrate on real chemistry how severe and how invisible it
becomes under imbalance, to explain it quantitatively, and to lay out a
practical protocol that restores per-class reliability.
\end{abstract}

\vspace{0.5em}

\section{Introduction}

A screening model is only useful if you can trust the cases where it says it is
sure. Modern classifiers for molecular property prediction reach high aggregate
accuracy, but a single accuracy number says nothing about \emph{which}
predictions deserve confidence. That gap has pushed conformal prediction into
cheminformatics. Conformal methods wrap an existing classifier and return, for
each compound, a set of labels that provably contains the truth at a
user-chosen rate~\citep{vovk2005algorithmic, angelopoulos2021gentle}. The
guarantee is distribution-free and finite-sample. It needs no assumption about
the model being correct, only that calibration and test compounds are
exchangeable. For a field that routinely commits expensive wet-lab resources on
the strength of a model's ranking, a guarantee like that is worth a great deal.

There is a catch that is easy to read past. The standard guarantee is
\emph{marginal}: coverage holds on average over the whole test distribution. It
does not promise anything about any particular subgroup. On a balanced problem
the distinction rarely bites. Screening problems are almost never balanced. The
interesting label---a toxic compound, a non-penetrant candidate, an active
hit---is usually the rare one, and the rare one is the whole reason for running
the campaign. A method that keeps its overall promise by being nearly perfect on
the common class and quietly poor on the rare class is, for screening, keeping
the wrong promise.

This paper measures how badly that plays out, and shows it is not a corner case.
On three imbalanced MoleculeNet tasks, marginal conformal prediction meets its
$90\%$ target overall while delivering minority-class coverage as low as
$38.9\%$. We then ask whether the effect is an artifact of one model family. It
is not. The same collapse appears with fingerprint-based random forests, with a
graph neural network, and with a pretrained transformer that treats a molecule as
a string of text---three ways of representing a molecule that share almost nothing
mechanically. The severity of the collapse turns out to depend on how
well-calibrated the base model already is on rare labels, not on which
representation it uses. That is a statement about the calibration step, not about
deep learning.

What makes the failure dangerous in practice is that it is invisible from the top.
Overall coverage, overall accuracy, ROC-AUC---all of the dashboard numbers a team
would normally check---stay healthy while minority coverage falls through the
floor. A practitioner who trusts the conformal guarantee, as advertised, would
have no reason to look closer.

The fix is not new and we do not claim it as ours. Class-conditional, or
Mondrian, conformal prediction calibrates a separate threshold per
class~\citep{vovk2013conditional}, which restores coverage within each group by
construction. The marginal-versus-conditional gap is textbook conformal
theory~\citep{angelopoulos2021gentle}. What we contribute is the empirical and
explanatory case that this textbook gap is a live hazard in virtual screening: we
quantify it on four datasets including one held out from the rest of our analysis, show it is
consistent across three representative molecular-learning paradigms, explain its size with a simple decomposition, trace it to specific
chemical substructures, and tie it to a decision-cost model that makes the practical
stakes concrete. Our aim is less ``here is a method'' and more ``here is a
vulnerability, here is why it happens, here is how to see it, and here is how to
close it.''

Concretely, the paper makes five claims, each backed by experiment:
\begin{enumerate}
\item On imbalanced molecular datasets, marginal conformal prediction satisfies
its global coverage target while under-covering the minority class, and the
shortfall grows with imbalance, from a $0.6$-point gap on near-balanced BACE to a
near-total collapse ($4.2\%$ coverage) on heavily imbalanced ClinTox.
\item The collapse reproduces across three unrelated model architectures at
$p<0.001$ each, with severity governed by baseline minority calibration rather
than architecture.
\item A simple decomposition of coverage explains the effect: the minority shortfall
equals the majority surplus amplified by the imbalance ratio. It follows directly from
the definition of weighted coverage, yet it predicts the measured gaps and their
ordering across datasets.
\item Standard aggregate metrics conceal the failure, which we demonstrate with a
selective-prediction protocol and a cost analysis, and the failure persists under
realistic scaffold splits.
\item The confidently wrong minority predictions concentrate on a small set of
common, low-information scaffolds, which we present as a chemical case study.
\end{enumerate}

\section{Related Work}

\paragraph{Conformal prediction and its conditional variants.}
The framework originates with Vovk and
colleagues~\citep{vovk2005algorithmic}, who established the exchangeability-based
coverage guarantee. The accessible modern treatment by
Angelopoulos and Bates~\citep{angelopoulos2021gentle} is now the standard reference, and it states
plainly that the basic guarantee is marginal and that conditional coverage
requires extra machinery. Two score functions matter for classification. The
least-ambiguous-set classifier (LAC) of Sadinle et al.~\citep{sadinle2019least} uses
$1-\hat{p}(y\mid x)$ and yields the smallest sets at a given coverage; adaptive
prediction sets~\citep{romano2020classification} trade size for better
conditional behavior. Class-conditional calibration, the Mondrian taxonomy of
Vovk~\citep{vovk2013conditional}, splits the calibration set by label and computes a
separate quantile for each, which gives coverage \emph{per class} at the price of
larger sets where a class is hard. Our work does not invent any of this. We use
LAC because it is the common default and because its small sets make the marginal
failure starkest, and we use Mondrian calibration as the remedy.

\paragraph{Conformal prediction in cheminformatics.}
Conformal methods entered drug discovery largely through the work of Norinder,
Carlsson, and colleagues, who reframed the classical applicability-domain question
as a coverage question~\citep{norinder2014introducing}, and through reviews aimed
at the discovery
pipeline~\citep{eklund2015application}. That line of work has mostly emphasized
what conformal prediction \emph{provides}---valid sets, an interpretable
confidence---rather than where the default formulation breaks. The imbalance
problem is acknowledged in passing in the general literature but, to our reading,
has not been measured systematically on standard molecular benchmarks, across
architectures, or connected to decision cost. That is the space this paper
occupies.

\paragraph{Calibration is a separate axis.}
It is tempting to assume that a miscalibrated model is the cause and a calibration
fix is the cure. Guo et al.~\citep{guo2017calibration} showed that modern networks are often
overconfident, and post-hoc scaling helps. We find the conformal failure is only
loosely related to ordinary calibration error: our base models are at worst
moderately miscalibrated (expected calibration error around $0.04$--$0.10$), yet
marginal conformal coverage still collapses on the minority. Calibration and
conditional coverage are different properties, and fixing the first does not fix
the second.

\paragraph{Molecular representations.}
We deliberately span three representation families. Circular (ECFP-style)
fingerprints~\citep{rogers2010extended} feed a random
forest~\citep{breiman2001random}. A graph convolutional
network~\citep{kipf2017semi} built with PyTorch
Geometric~\citep{fey2019fast} learns from the atom-bond graph. ChemBERTa, a RoBERTa-style
transformer pretrained on millions of SMILES strings~\citep{chithrananda2020chemberta},
supplies sequence embeddings. These have essentially nothing in common at the
level of how a molecule is encoded, which is what makes their shared failure
informative.

\section{Methods}

\subsection{Datasets}
We use four binary classification tasks from
MoleculeNet~\citep{wu2018moleculenet}, chosen to span a wide range of class imbalance
(Table~\ref{tab:datasets}). Blood--brain-barrier penetration
(BBBP)~\citep{martins2012bayesian} labels whether a compound crosses the barrier;
the minority class is the non-penetrant compounds. BACE~\citep{subramanian2016computational}
labels $\beta$-secretase~1 inhibition and is close to balanced. From
Tox21~\citep{huang2016tox21} we take the SR-ARE stress-response endpoint. ClinTox
provides the CT-TOX endpoint (failure of clinical trials for toxicity); it is the most
imbalanced task we use. Like the others it is a MoleculeNet task, so it serves as a
held-out check---unused anywhere else in our study---rather than independent external
validation from a separate data source. SMILES strings that RDKit~\citep{landrum_rdkit}
could not parse were dropped; the counts in Table~\ref{tab:datasets} are
post-filtering and are identical across all experiments on a given dataset.

Most experiments use random train/calibration/test splits, which preserve the
exchangeability conformal prediction assumes. For the robustness check in
Section~\ref{sec:scaffold-split} we instead use scaffold splits: molecules are grouped
by Bemis--Murcko scaffold and whole groups are assigned to a single partition, so that
test compounds carry skeletons unseen in training and calibration. Scaffold splits
break exchangeability by design and so test behavior under realistic distribution
shift.

\begin{table}[t]
\centering
\caption{Datasets after parsing. ``Minority rate'' is the fraction of compounds
in the smaller class.}
\label{tab:datasets}
\begin{tabular}{lrrrl}
\toprule
Dataset & Compounds & Class 0 & Class 1 & Minority rate \\
\midrule
BACE          & 1513 & 822  & 691  & 0.46 \\
BBBP          & 2039 & 479  & 1560 & 0.23 \\
Tox21 SR-ARE  & 5825 & 4883 & 942  & 0.16 \\
ClinTox CT-TOX & 1480 & 1368 & 112 & 0.08 \\
\bottomrule
\end{tabular}
\end{table}

\subsection{Models and representations}
For the main conformal experiments and all cross-dataset work we use a random
forest~\citep{breiman2001random} (500 trees) on $2048$-bit Morgan fingerprints of
radius~2, computed with RDKit's \texttt{MorganGenerator}. We selected the random
forest after comparing four standard classifiers on BBBP under stratified
five-fold cross-validation (Section~\ref{sec:base}); it gave the best AUC and
competitive balanced accuracy, and as a bagged ensemble it produces reasonable
probability estimates out of the box.

For the architecture study we add two more representations. The graph network is a
two-layer graph convolutional model~\citep{kipf2017semi}
($\text{GCNConv}\,32{\to}64{\to}64$, mean pooling, a linear read-out to two
classes) implemented in PyTorch Geometric~\citep{fey2019fast}, trained for 60
epochs with Adam at learning rate $10^{-3}$. Atom features are a 32-dimensional
one-hot encoding of element, degree, hydrogen count, hybridization, aromaticity,
and formal charge. The transformer is ChemBERTa
(\texttt{DeepChem/ChemBERTa-77M-MTR})~\citep{chithrananda2020chemberta}, used as a
frozen feature extractor: we mean-pool its final hidden states over the SMILES
tokens to a 384-dimensional embedding and fit a logistic-regression head. Freezing
the transformer isolates the representation from any tuning choices on our side.

\subsection{Conformal prediction}
We use split (inductive) conformal prediction. Each dataset is partitioned into a
proper training set ($50\%$), a calibration set ($25\%$), and a test set
($25\%$). The base model is fit on the training split only. For a model output
$\hat{p}(y\mid x)$ we score each calibration point with the LAC nonconformity
score~\citep{sadinle2019least}
\begin{equation}
s_i \;=\; 1 - \hat{p}(y_i \mid x_i).
\end{equation}
For target coverage $1-\alpha$ with $\alpha=0.10$, the marginal threshold is the
conformal quantile
\begin{equation}
\hat{q} \;=\; \text{the } \big\lceil (n+1)(1-\alpha) \big\rceil / n \text{ empirical quantile of } \{s_i\},
\label{eq:quantile}
\end{equation}
where $n$ is the calibration-set size, and the prediction set for a test point is
\begin{equation}
C(x) \;=\; \{\, y : \hat{p}(y\mid x) \ge 1-\hat{q} \,\}.
\end{equation}
The finite-sample correction in Eq.~\eqref{eq:quantile} is what gives the marginal
guarantee $\Pr\!\big(y \in C(x)\big) \ge 1-\alpha$.

Class-conditional (Mondrian) calibration changes one thing. Instead of a single
threshold it computes a separate quantile $\hat{q}_k$ from the calibration points
of each class $k$:
\begin{equation}
\hat{q}_k \;=\; \text{the conformal quantile of } \{\, s_i : y_i = k \,\},
\qquad
C(x) \;=\; \{\, y : \hat{p}(y\mid x) \ge 1-\hat{q}_y \,\}.
\end{equation}
This enforces $\Pr\!\big(y \in C(x) \mid y=k\big) \ge 1-\alpha$ for every class,
which is precisely the per-class guarantee the marginal version lacks.

We report coverage overall and within each class, and mean prediction-set size.
Every conformal experiment is repeated over ten random
train/calibration/test splits (seeds $0$--$9$); we report the mean and, where
relevant, the standard deviation across seeds.

\subsection{Selective prediction}
A conformal classifier can also be read as an accept/abstain rule. A singleton
prediction set is a confident, actionable call; a two-element set $\{0,1\}$ is an
abstention---the model declines and, in a screening setting, the compound goes to
the lab. We report the acceptance rate (fraction of singletons) and the accuracy
on accepted compounds, overall and for the minority class, under both calibration
schemes.

\subsection{Decision-cost model}
To turn coverage into something a project manager would recognize, we score
minority-class decisions with a simple utility
\begin{equation}
U \;=\; B_{\text{tp}} N_{\text{tp}} \;-\; C_{\text{fp}} N_{\text{fp}} \;-\; C_{\text{fn}} N_{\text{fn}} \;-\; C_{\text{lab}} N_{\text{sent}},
\label{eq:utility}
\end{equation}
where $N_{\text{tp}}$, $N_{\text{fp}}$, $N_{\text{fn}}$ count correct and
incorrect confident calls, $N_{\text{sent}}$ counts abstentions routed to the lab,
and $B,C$ are the corresponding benefit and costs. The point is not the exact
numbers but the structure: a wrong confident call is far more expensive than a lab
assay. We use $B_{\text{tp}}=1$, $C_{\text{fp}}=C_{\text{fn}}=5$,
$C_{\text{lab}}=0.5$ over $1000$ minority decisions, plugging in the measured
acceptance rates and selective accuracies.

\subsection{Scaffold blind-spot analysis}
To find \emph{where} the marginal model is confidently wrong, we reduce each
compound to its Bemis--Murcko scaffold~\citep{bemis1996properties}---its ring and
linker skeleton, side chains removed---using RDKit. On the BBBP test split we flag
every minority compound whose marginal prediction set is a confident singleton
that excludes the true (minority) label. We then count these confident errors per
scaffold and rank the scaffolds with at least two minority test compounds.

\subsection{Statistical testing}
For each architecture we compare minority coverage under Mondrian versus marginal
calibration with a one-sided Wilcoxon signed-rank test over the ten paired seeds.
We report $95\%$ confidence intervals for minority coverage by percentile bootstrap
($10{,}000$ resamples over the ten seeds). To check that the failure is not specific
to the LAC score, we also evaluate randomized adaptive prediction sets
(APS)~\citep{romano2020classification}, calibrated both marginally and per class.
We also report expected calibration error
(ECE)~\citep{guo2017calibration} and Brier scores for the base models.

\section{Results}

\subsection{Base classifier selection and calibration}
\label{sec:base}
Table~\ref{tab:base} compares four classifiers on BBBP. The random forest leads on
AUC ($0.916$) with a histogram gradient-boosting model close behind and slightly
ahead on balanced accuracy. We carried the random forest forward as the base model
for all conformal experiments.

\begin{table}[t]
\centering
\caption{Base classifiers on BBBP under stratified five-fold cross-validation
(mean over folds). ROC-AUC reported with its standard deviation.}
\label{tab:base}
\begin{tabular}{lccc}
\toprule
Classifier & ROC-AUC & Balanced acc. & $F_1$ \\
\midrule
Random forest          & $0.916 \pm 0.007$ & 0.794 & 0.931 \\
Hist.\ gradient boost  & $0.912 \pm 0.009$ & \textbf{0.816} & 0.928 \\
SVM (RBF)              & $0.891 \pm 0.013$ & 0.766 & 0.927 \\
Logistic regression    & $0.881 \pm 0.010$ & 0.788 & 0.913 \\
\bottomrule
\end{tabular}
\end{table}

\begin{figure}[t]
\centering
\includegraphics[width=\textwidth]{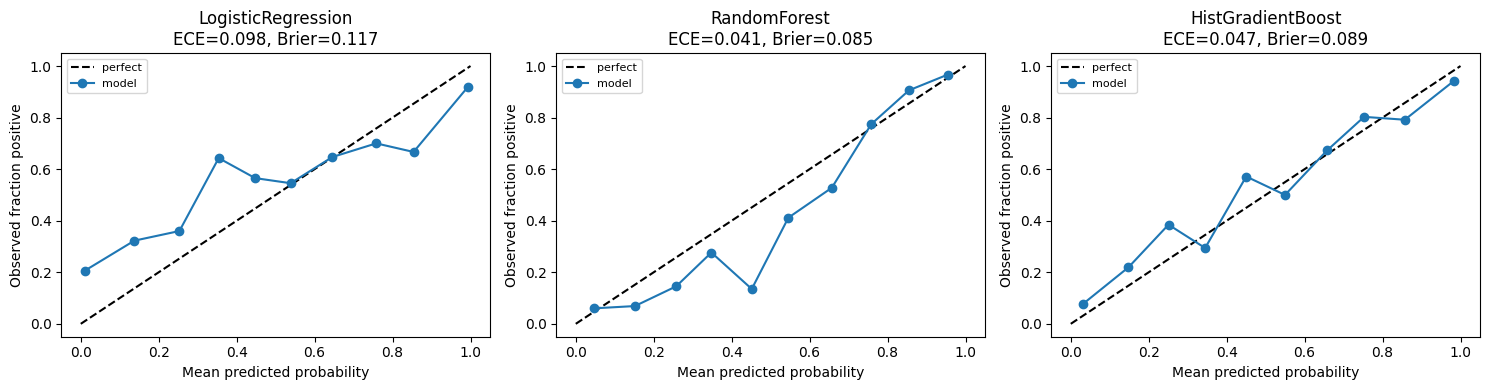}
\caption{Reliability diagrams for three base classifiers on BBBP (out-of-fold).
Deviation from the diagonal is mild; expected calibration error stays between
$0.04$ and $0.10$. Ordinary calibration being acceptable is what makes the
conformal coverage failure in Figure~\ref{fig:coverage} surprising.}
\label{fig:calibration}
\end{figure}

The base models are not badly calibrated. Out-of-fold expected calibration error
ranges from $0.041$ (random forest) to $0.098$ (logistic regression), with Brier
scores between $0.085$ and $0.117$; the reliability diagrams in
Figure~\ref{fig:calibration} show only mild deviations from the diagonal, mostly
underconfidence at low predicted probabilities for the linear model. We make a
point of this because it rules out the obvious explanation. The conformal failure
below is not a symptom of a wildly overconfident model. It survives even when
ordinary calibration is fine.

\subsection{Marginal coverage collapses on the minority class}
\begin{figure}[t]
\centering
\includegraphics[width=0.62\textwidth]{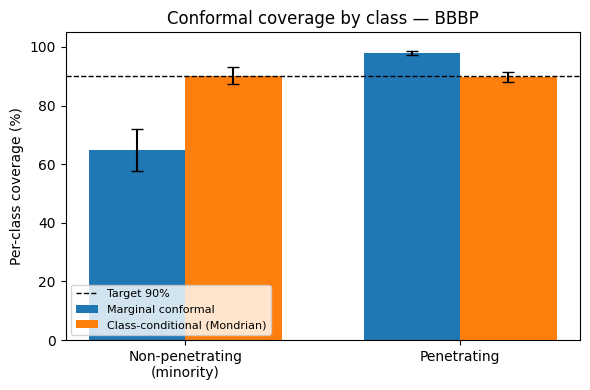}
\caption{Per-class coverage on BBBP at a $90\%$ target (random forest, ten seeds).
Marginal calibration over-covers the majority and leaves the minority at
$64.8\%$. Mondrian calibration brings both classes to target.}
\label{fig:coverage}
\end{figure}

At a $90\%$ target, marginal conformal prediction on BBBP behaves exactly as
advertised \emph{in aggregate}: overall coverage is $90.2\%$. Split that number by
class and it falls apart. Majority-class coverage is $98.1\%$; minority-class
coverage is $64.8\%$ ($\pm 7.0$ across seeds; $95\%$ bootstrap CI $[59.8, 67.8]$). The method buys its global
guarantee by over-covering the easy class and under-covering the hard one
(Figure~\ref{fig:coverage}). A practitioner watching overall coverage would see a
$90.2\%$ that looks like the guarantee working. The compounds they care about are
covered two times in three.

Switching to Mondrian calibration repairs this. Overall coverage stays on target
($89.8\%$), majority coverage drops to a no-longer-inflated $89.6\%$, and minority
coverage rises to $90.3\%$ ($\pm 3.0$)---on target, and with lower seed-to-seed
variance. The cost is set size: mean prediction-set size grows from $1.05$ to
$1.26$. In other words the marginal method was producing tight sets partly by
omitting the true minority label; the conditional method pays for honest coverage
with slightly larger sets on the hard class, which is the correct trade.

\subsection{Severity scales with imbalance}
\label{sec:severity}
\begin{figure}[t]
\centering
\includegraphics[width=0.72\textwidth]{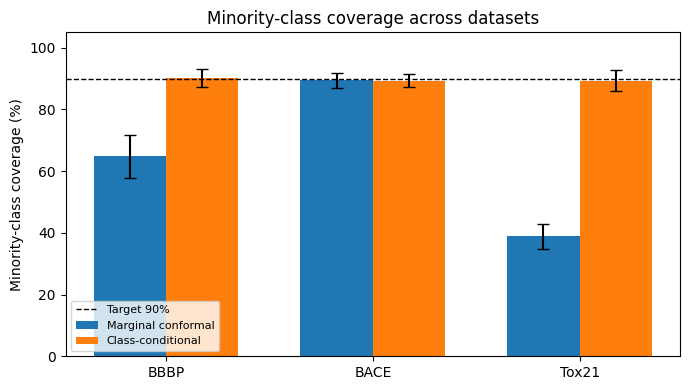}
\caption{Marginal minority-class coverage against a $90\%$ target (dashed) on three
datasets ordered by imbalance. The shortfall deepens from BACE to BBBP to Tox21;
Mondrian calibration returns every dataset to target.}
\label{fig:generalization}
\end{figure}

The same experiment across datasets lines up with their imbalance
(Table~\ref{tab:generalization}, Figure~\ref{fig:generalization}). On near-balanced
BACE (minority rate $0.46$) marginal calibration is already fine: minority coverage
is $89.4\%$, and Mondrian leaves it essentially unchanged at $89.3\%$ with no
change in set size. On BBBP ($0.23$) the gap opens to the $64.8\%$ above. On Tox21
SR-ARE ($0.16$) it becomes severe: marginal minority coverage is $38.9\%$ against
the $90\%$ target, a shortfall of more than fifty points. Mondrian restores it to
$89.4\%$, here at a larger set-size cost ($1.12 \to 1.49$) because the minority
class on Tox21 is genuinely harder to cover. The ordering is monotone: the rarer
the class, the worse marginal calibration treats it, and BACE shows the effect
switches off when imbalance does.

To name the failure with a single number, we use the \emph{minority coverage gap},
$\mathrm{MCG} = (1-\alpha) - \mathrm{cov}_{\mathrm{min}}$: how far minority coverage
falls below the promised target. A companion, the \emph{coverage imbalance index}
$\mathrm{CII} = |\mathrm{cov}_{\mathrm{maj}} - \mathrm{cov}_{\mathrm{min}}|$, captures
how unevenly coverage is split between classes. These are deliberately simple; the
value of a diagnostic is that it is easy to compute and report, not that it is
mathematically deep. Under marginal calibration $\mathrm{MCG}$ is $0.6$ (BACE),
$25.2$ (BBBP), and $51.1$ (Tox21) points; under Mondrian it collapses to within a
point of zero on all three. On BBBP, where we can resolve both classes,
$\mathrm{CII}$ falls from $33.3$ points (marginal) to $0.7$ (Mondrian).

We also held out a fourth MoleculeNet task, ClinTox, used nowhere else in our
experiments. It labels compounds by clinical-trial toxicity failure; its toxic class is the
rarest minority we study (rate $0.08$). Here the collapse is near-total: marginal
minority coverage is $4.2\%$ against the $90\%$ target, an $\mathrm{MCG}$ of nearly
$86$ points. Mondrian recovers it to $94.6\%$ (the slight over-shoot reflects the
small minority test set, roughly $28$ compounds per split, which makes coverage
estimates coarse). Every dataset, including this held-out one, follows the same
pattern, and Section~\ref{sec:mechanism} shows why the ordering is not a
coincidence.

\begin{table}[t]
\centering
\caption{Minority-class coverage at a $90\%$ target across datasets (random forest,
ten seeds). The marginal shortfall grows with the imbalance amplification factor
$\pi_{\mathrm{maj}}/\pi_{\mathrm{min}}$ (Section~\ref{sec:mechanism}). ClinTox is a
fourth MoleculeNet task, held out from the rest of our analysis.}
\label{tab:generalization}
\begin{tabular}{lccccc}
\toprule
Dataset & Minority rate & $\pi_{\mathrm{maj}}/\pi_{\mathrm{min}}$ & Marginal cov. & Mondrian cov. & Set size (marg.\,$\to$\,mond.) \\
\midrule
BACE          & 0.46 & 1.19 & 89.4\% & 89.3\% & $1.23 \to 1.23$ \\
BBBP          & 0.23 & 3.26 & 64.8\% & 90.3\% & $1.05 \to 1.26$ \\
Tox21 SR-ARE  & 0.16 & 5.18 & 38.9\% & 89.4\% & $1.12 \to 1.49$ \\
ClinTox CT-TOX & 0.08 & 12.2 & \phantom{0}4.2\% & 94.6\% & $0.99 \to 1.75$ \\
\bottomrule
\end{tabular}
\end{table}

\subsection{The failure persists across model paradigms}
If the collapse were specific to fingerprints and random forests, it would be a
quirk worth a footnote. It is not. We repeated the BBBP experiment with two more
representations that share nothing with the first: a graph network reading the
molecular graph, and a frozen chemical language model reading the SMILES string.
Both collapse the same way (Table~\ref{tab:architecture}). Marginal minority
coverage is $63.3\%$ for the GNN ($95\%$ bootstrap CI $[60.6, 65.8]$) and $72.4\%$
for the transformer ($[69.5, 75.0]$), against the
$90\%$ target; Mondrian restores $90.9\%$ and $90.1\%$ respectively. The Wilcoxon
test is at its floor for ten paired seeds in all three cases, $p=9.8\times10^{-4}$,
with every seed pointing the same direction.

\begin{table}[t]
\centering
\caption{Minority-class coverage on BBBP at a $90\%$ target across three model
architectures (ten seeds each). The failure and its repair appear in all three.}
\label{tab:architecture}
\begin{tabular}{llccc}
\toprule
Architecture & Representation & Marginal cov. & Mondrian cov. & Wilcoxon $p$ \\
\midrule
Random forest & ECFP fingerprints     & 64.8\% & 90.3\% & $9.8\times10^{-4}$ \\
Graph network & atom--bond graph       & 63.3\% & 90.9\% & $9.8\times10^{-4}$ \\
ChemBERTa     & SMILES (frozen)        & 72.4\% & 90.1\% & $9.8\times10^{-4}$ \\
\bottomrule
\end{tabular}
\end{table}

One detail in Table~\ref{tab:architecture} is worth dwelling on. The transformer's
marginal coverage is noticeably higher than the other two, $72.4\%$ versus around
$64\%$. ChemBERTa was pretrained on millions of molecules, so it arrives with
better-calibrated minority predictions, and the marginal threshold therefore
mistreats the minority a little less. But ``a little less'' is still an $18$-point
miss. The vulnerability shrinks with a stronger pretrained model; it does not
disappear. We read this as evidence that the collapse is driven by the marginal
calibration step interacting with baseline minority calibration, not by anything
architectural---a better representation raises the floor without removing the
problem.

\subsection{The failure is not specific to one conformal score}
\label{sec:aps}
A second natural objection is that we picked a nonconformity score (LAC) that happens
to fail. To check, we repeated the BBBP experiment with adaptive prediction sets
(APS)~\citep{romano2020classification}, a different and widely used score, in its
randomized form. The pattern is the same (Table~\ref{tab:aps}). Under marginal
calibration APS under-covers the minority at $74.6\%$ (95\% bootstrap CI
$[71.8, 77.6]$)---less severely than LAC's $64.1\%$, because APS builds slightly
larger sets, but still far below the $90\%$ target. Class-conditional calibration
restores it to $89.1\%$ for APS, just as it does for LAC. The two scores trade size
for coverage differently, yet both collapse on the minority under marginal
calibration and both are repaired by Mondrian. The failure is a property of pooling
the calibration set under imbalance, not of the particular score.

\begin{table}[t]
\centering
\caption{Two nonconformity scores on BBBP minority coverage at a $90\%$ target
(random forest, ten seeds, $95\%$ bootstrap CIs). Both under-cover the minority under
marginal calibration and both are restored by Mondrian.}
\label{tab:aps}
\begin{tabular}{llcc}
\toprule
Score & Calibration & Minority cov.\ [95\% CI] & Set size \\
\midrule
LAC & marginal & 64.1\% $[59.8, 67.8]$ & 1.06 \\
LAC & Mondrian & 90.0\% $[87.6, 92.4]$ & 1.23 \\
APS & marginal & 74.6\% $[71.8, 77.6]$ & 1.16 \\
APS & Mondrian & 89.1\% $[85.9, 92.5]$ & 1.44 \\
\bottomrule
\end{tabular}
\end{table}

\subsection{Aggregate metrics hide the failure}
The selective-prediction view on BBBP shows why a normal evaluation would miss all
of this. Under marginal calibration the model accepts (returns a confident
singleton for) $95\%$ of compounds at $89.7\%$ accuracy on those accepted---an
excellent-looking summary. Restrict to the minority class and the same model
accepts $84\%$ of compounds at $58.4\%$ selective accuracy. It is confidently
wrong on minority compounds close to half the time, while the headline accuracy on
accepted compounds reads $89.7\%$. Mondrian calibration changes the behavior in the
right direction: it accepts fewer minority compounds when it should be unsure
($87\%$ acceptance is reached only because the larger sets push more genuinely
uncertain cases to abstention) and lifts minority selective accuracy to $88.8\%$.
The overall accept/accuracy figures barely move between the two schemes. That is
the trap. Everything you would normally look at says the marginal model is fine.

\subsection{The cost of false confidence}
Feeding the measured minority numbers into the utility of Eq.~\eqref{eq:utility}
makes the stakes concrete. Over $1000$ minority decisions, a forced classifier that
never abstains scores $U=-1496$: at $58.4\%$ accuracy on a problem where wrong
calls cost five times a lab assay, confident automation loses money. Marginal
conformal prediction, abstaining a little, improves this only to $U=-1336.6$---it
abstains on the wrong compounds, because its confident singletons are exactly where
it is overconfident about the minority. Mondrian calibration scores $U=+220.4$, a
swing of $+1557$ relative to marginal. The qualitative result---abstention is only
valuable if you abstain on the right cases---is robust to the exact cost
constants; with any cost structure where a confident error outweighs a lab assay,
conditional calibration is what makes selective prediction pay.

\begin{figure}[t]
\centering
\includegraphics[width=0.85\textwidth]{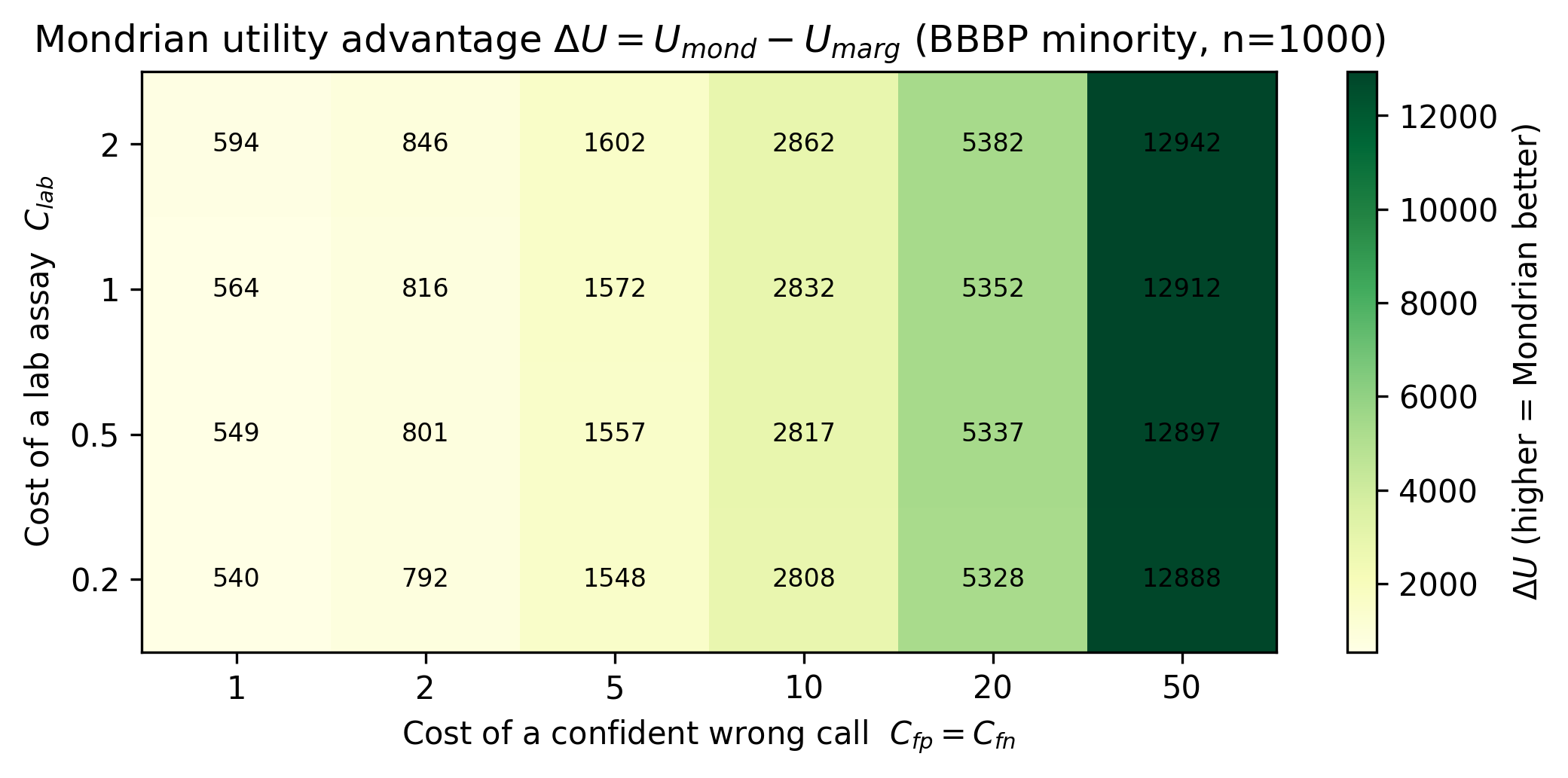}
\caption{Sensitivity of the Mondrian utility advantage $\Delta U = U_{\mathrm{mond}}
- U_{\mathrm{marg}}$ to the cost assumptions, on BBBP minority decisions ($n=1000$).
$\Delta U$ is positive across every combination of error cost and lab-assay cost,
so class-conditional calibration is the better decision rule regardless of where a
campaign sets its costs.}
\label{fig:cost}
\end{figure}

To make ``robust to the exact constants'' more than an assertion, we swept the cost
of a confident wrong call ($C_{\mathrm{fp}}=C_{\mathrm{fn}}$) from $1$ to $50$ and
the cost of a lab assay ($C_{\mathrm{lab}}$) from $0.2$ to $2$
(Figure~\ref{fig:cost}). The Mondrian advantage $\Delta U = U_{\mathrm{mond}} -
U_{\mathrm{marg}}$ is positive everywhere on the grid, ranging from $+540$ to
$+12{,}942$ over $1000$ minority decisions, and it grows as confident errors get
more expensive---which is the regime any safety-critical screen lives in. The
conclusion does not depend on our particular choice of $5$ and $0.5$.

\subsection{Where the model is blind: a scaffold case study}
\begin{figure}[t]
\centering
\includegraphics[width=\textwidth]{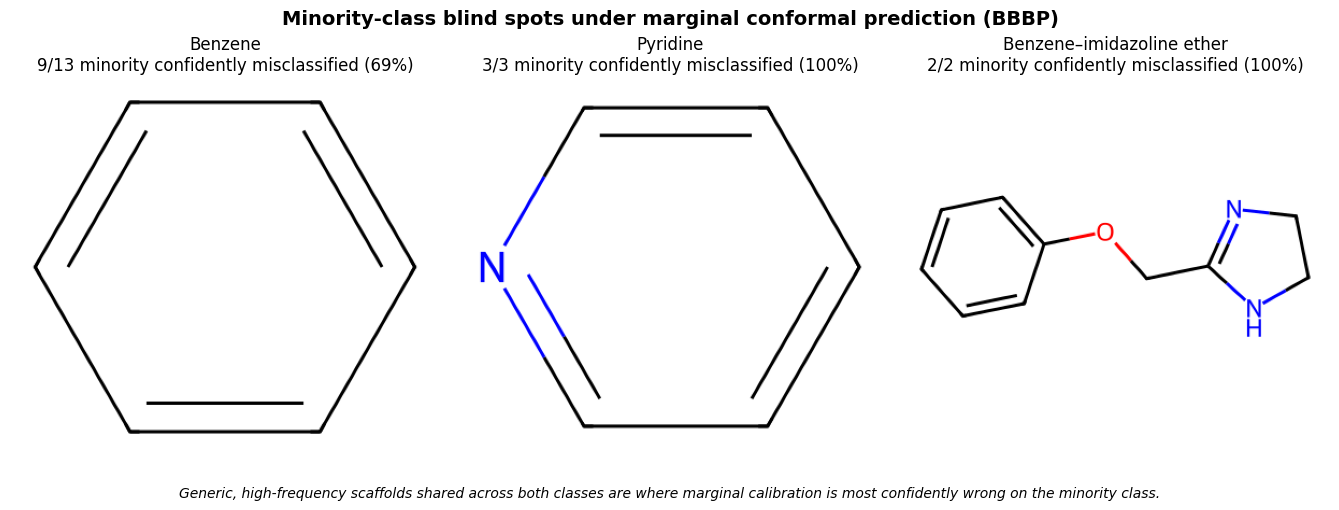}
\caption{The three Bemis--Murcko scaffolds with the most confident minority-class
errors under marginal calibration on BBBP. The worst blind spots are the most
generic cores (benzene, pyridine), which appear in both classes and carry little
class signal.}
\label{fig:scaffold}
\end{figure}

The confident minority errors are not spread evenly across chemical space. They
pile up on a few scaffolds (Figure~\ref{fig:scaffold}). On the BBBP test split the
worst offender is the bare benzene scaffold: of $13$ minority compounds reducing to
a benzene core, the marginal model is confidently wrong on $9$ ($69\%$). Pyridine
is next ($3$ of $3$), followed by a benzene--imidazoline ether ($2$ of $2$). The
common thread is telling. These are the simplest, most frequent skeletons in the
dataset, and they occur in \emph{both} classes. A bare benzene ring carries almost
no information about whether a compound crosses the blood--brain barrier, so the
model defaults to the majority call and does it confidently---and it is exactly the
minority compounds hiding inside that generic chemistry that get burned. Complex,
distinctive scaffolds (fused polycyclic systems) produced zero confident minority
errors. The blind spot is not exotic chemistry the model never saw. It is the most
ordinary chemistry, where the minority signal is drowned out. That is a sharper and
more uncomfortable statement than ``the model fails on weird molecules,'' and it is
precisely the regime a marginal guarantee papers over.

\subsection{Robustness to realistic scaffold splits}
\label{sec:scaffold-split}
Random train/test splits let near-identical molecules land on both sides, which
flatters any model and, in principle, could flatter the conformal calibration too.
The honest test is a scaffold split, where whole Bemis--Murcko scaffolds are assigned
to a single partition so that test compounds carry skeletons absent from training and
calibration. This deliberately violates the exchangeability that conformal coverage
assumes, so it probes robustness under distribution shift rather than a clean
guarantee. We repeated the BBBP experiment with ten random scaffold splits. The
collapse does not go away; it deepens. Marginal minority coverage falls to
$61.3\%$ (from $64.8\%$ under random splits), and Mondrian recovers it to $94.1\%$
(Wilcoxon $p=9.8\times10^{-4}$, all ten splits in agreement). The per-split marginal
numbers are noisy---each scaffold partition is a different shift, so they range from
$38\%$ to $80\%$---and Mondrian over-covers slightly rather than landing exactly on
target, which is the expected and safe direction when exchangeability is broken. The
takeaway is that the failure is not an artifact of optimistic splitting: on the
evaluation that mirrors deployment to new chemistry, marginal calibration is if
anything worse, and class-conditional calibration still repairs it.

\section{Mechanism: why marginal calibration under-covers the minority}
\label{sec:mechanism}

The experiments establish that the failure is real, general, and costly. They do not
yet say why. Here we give a simple decomposition of coverage that explains the effect,
predicts its size from quantities known before any model is trained, and accounts for
the ordering across datasets. It follows immediately from the definition of weighted
coverage; we present it not as a new theoretical result but as an explanatory tool that
happens to be quantitatively accurate.

\begin{figure}[t]
\centering
\includegraphics[width=0.85\textwidth]{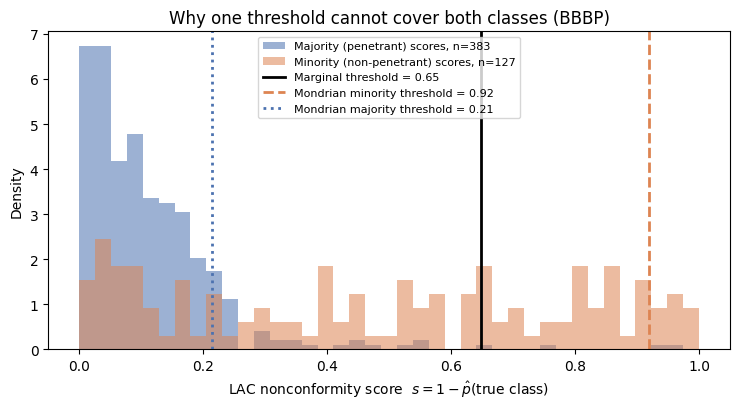}
\caption{Calibration nonconformity scores on BBBP, split by class. The majority
(penetrant) scores concentrate near zero; the minority (non-penetrant) scores spread
to high values. The single marginal threshold is pulled toward the majority mass and
slices through the minority distribution, excluding everything to its right---the
source of the under-coverage. Mondrian gives each class its own threshold; the minority
threshold sits far enough right to cover its tail.}
\label{fig:mechanism}
\end{figure}

\paragraph{Conservation of coverage.}
Split conformal prediction guarantees only the \emph{overall} coverage,
$\Pr(y \in C(x)) \ge 1-\alpha$. Overall coverage is the class-weighted average of the
per-class coverages, with weights equal to the class prevalences
$\pi_{\mathrm{maj}}, \pi_{\mathrm{min}}$:
\begin{equation}
1-\alpha \;=\; \pi_{\mathrm{maj}}\,\mathrm{cov}_{\mathrm{maj}}
\;+\; \pi_{\mathrm{min}}\,\mathrm{cov}_{\mathrm{min}}.
\label{eq:conservation}
\end{equation}
Solving for the minority coverage and rearranging gives the relation that drives the
whole phenomenon:
\begin{equation}
\underbrace{(1-\alpha) - \mathrm{cov}_{\mathrm{min}}}_{\text{minority shortfall}}
\;=\;
\frac{\pi_{\mathrm{maj}}}{\pi_{\mathrm{min}}}\;
\underbrace{\big(\mathrm{cov}_{\mathrm{maj}} - (1-\alpha)\big)}_{\text{majority surplus}}.
\label{eq:amplification}
\end{equation}
Coverage is conserved: whatever the majority is given above target, the minority must
lose below it. And the loss is not one-for-one---it is amplified by the imbalance
ratio $\pi_{\mathrm{maj}}/\pi_{\mathrm{min}}$. At three-to-one imbalance, each point
of majority over-coverage costs three points of minority coverage. The mechanism is
visible directly in the calibration scores (Figure~\ref{fig:mechanism}): because the
model classifies the abundant majority easily, its nonconformity scores pile up near
zero, so the pooled $(1-\alpha)$ quantile---the single marginal threshold---is dragged
down to a value that comfortably covers the majority while failing to reach the
minority's much heavier upper tail. The majority's surplus is exactly the over-coverage
that this low threshold produces.

\paragraph{The identity predicts the failure.}
Equation~\eqref{eq:amplification} is not a loose analogy; it holds numerically. On
BBBP the majority's over-coverage ($\mathrm{cov}_{\mathrm{maj}}-(1-\alpha)$) is $8.1$ points
and the imbalance ratio ($\pi_{\mathrm{maj}}/\pi_{\mathrm{min}}$) is $3.26$,
so the identity predicts a minority coverage of $63.6\%$. The experiment measured
$64.8\%$---a gap of just over one point, the rest accounted for by finite-sample
quantile corrections we suppressed for clarity. The amplification factor also explains
why the datasets fall in the order they do. It climbs from $1.19$ (BACE) to $3.26$
(BBBP) to $5.18$ (Tox21) to $12.2$ (ClinTox), and the measured minority gaps climb in
lockstep: $0.6$, $25.2$, $51.1$, and $85.8$ points (Table~\ref{tab:generalization}).
Near balance the amplifier is close to one and the failure is negligible; at the
twelve-to-one imbalance of ClinTox it is large enough to erase minority coverage
almost entirely. Nothing about this is specific to chemistry: the only ingredients are
class imbalance and a model that handles the common class better than the rare one,
which is the normal situation in screening.

\paragraph{Why Mondrian works, and why temperature scaling would not.}
Mondrian calibration breaks Eq.~\eqref{eq:conservation} by refusing to pool. With a
separate quantile per class, each class gets its own $(1-\alpha)$ threshold and the
amplification channel is closed---there is no shared budget to redistribute. This also
clarifies why ordinary recalibration is the wrong tool. Rescaling probabilities (for
instance, temperature scaling) shifts every score but cannot change the fact that a
single pooled quantile is dominated by the majority; it adjusts confidence, not the
conditional coverage that Eq.~\eqref{eq:amplification} governs. The fix has to act on
the calibration partition, not on the probabilities.

\paragraph{An illusion of safety.}
Put the pieces together and a pattern emerges that is worse than any single number.
A team adopts conformal prediction precisely to stop trusting a model blindly. They
pick $\alpha=0.10$, confirm overall coverage near $90\%$, see high accuracy on
confident calls, and conclude the system is safe to act on. Every one of those
checks passes while the model covers the minority class two times in three, abstains
on the wrong compounds, and is most confidently wrong on the most ordinary
molecules in the library. The marginal guarantee did not lie---it delivered exactly
the average coverage it promised---but the way it is naturally read, as a per-case
assurance, is an illusion of safety on imbalanced data.

\paragraph{What is and is not new here.}
The mathematics of conformal prediction is not ours and we want to be exact about
that. The gap between marginal and conditional coverage is foundational conformal
theory~\citep{vovk2013conditional, angelopoulos2021gentle}, and Mondrian
calibration is the standard answer to it. Our contribution is empirical and
explanatory: that this known gap is large, common, and invisible under the imbalance
that defines virtual screening; that it holds across three unrelated architectures,
two nonconformity scores, and a held-out dataset, with severity set by baseline
minority calibration rather than architecture; that an elementary decomposition of
coverage predicts its size and
ordering from class prevalences alone; that ordinary metrics conceal it; and that it
localizes to interpretable chemical substructures with a measurable, cost-robust
penalty. The conservation argument in Section~\ref{sec:mechanism} is elementary, and
we make no claim otherwise---its worth is that it turns an observed failure into a
predictable one.

\paragraph{Practical recommendation.}
The operational takeaway is short. On any imbalanced screening task, report coverage
\emph{per class}, not just overall, and calibrate per class with Mondrian conformal
prediction by default. The extra cost is a modest increase in prediction-set size,
concentrated on the hard class, which is the honest place to spend it. Treating
larger minority sets as a defect is backwards; they are the method telling the truth
about a class it cannot yet call confidently.

\paragraph{Beyond virtual screening.}
Nothing in the mechanism is specific to molecules. The decomposition in
Section~\ref{sec:mechanism} needs only two ingredients: class imbalance, and a model
that handles the common class better than the rare one. That pairing is the rule, not
the exception, in the settings where calibrated abstention matters most---medical
diagnosis of rare conditions, fraud detection, rare-disease and adverse-event
prediction, and protein or sequence design where the desirable label is scarce. In
each, a marginal conformal wrapper would report a reassuring overall coverage while
quietly under-covering exactly the rare class the system exists to catch, and the
amplification factor says the under-coverage grows with the imbalance. We have not
tested these domains and make no empirical claim about them; we point them out because
the failure we measured in chemistry is a property of imbalanced calibration in
general, and per-class coverage reporting is a cheap safeguard wherever that situation
arises.

\paragraph{Limitations.}
Several remain. We study four datasets and one imbalance axis; a fuller sweep across
endpoints and ratios would turn the amplification relation from a validated trend
into a fitted law. Our scaffold-split test uses random scaffold partitions over ten
seeds rather than a single canonical split, which is why its coverage estimates are
noisier; the direction is nonetheless unambiguous. ClinTox is small, so its minority
test set is only about $28$ compounds per split and its percentages are coarse. We
froze ChemBERTa rather than fine-tuning it, so its numbers reflect pretrained
representations, not a tuned model. We report results at a single target $\alpha=0.10$;
we compared two nonconformity scores (LAC and APS) but did not sweep $\alpha$, and a
full sweep would confirm the effect is not specific to one error level. The cost model
is illustrative---we show the sign and robustness of the effect, not real campaign
economics. And the scaffold blind-spot analysis is run on one seed for
interpretability; the identity of the worst scaffolds may shift across splits even if
the generic-core pattern does not.

\paragraph{Future work.}
The natural extensions are a wider dataset sweep to fit the amplification relation
directly, a sweep over the target error level $\alpha$, and group-conditional
calibration along axes beyond the label---chemical series, assay source---where silent
under-coverage could hide for the same reason. A predictor that
estimates the coverage gap from prevalence and baseline calibration alone, turning the
conservation identity into a pre-calibration safety check, is a promising direction
once enough datasets are in hand to fit it credibly.

\section{Conclusion}

Conformal prediction gives virtual screening something it badly needs: a
distribution-free statement about reliability. Read carelessly on imbalanced data,
that statement is also a way to be confidently wrong about the compounds that matter
most. We measured marginal conformal prediction under-covering the minority class by
up to fifty points while every aggregate metric stayed healthy, showed the failure
across fingerprint, graph, and transformer models at $p<0.001$ each, traced it to a
handful of generic scaffolds, and priced it with a simple cost model. The remedy is
already on the shelf---calibrate per class. The point of the paper is to make the
case that, for screening, per-class calibration should be the default rather than a
refinement, and to give practitioners a concrete way to see the failure before it
costs them.

\section*{Data and Code Availability}
All datasets are public components of
MoleculeNet~\citep{wu2018moleculenet} (BBBP, BACE, Tox21). Code to reproduce every
experiment, table, and figure, with fixed random seeds, is available at
\url{https://github.com/Muhammadjon-crypto/Conformal-Minority-Screen}. A
Zenodo-archived release with a permanent DOI will be provided on acceptance.

\section*{Author Contributions}
Muhammadjon Tursunbadalov and Mustafojon Tursunbadalov jointly
designed the study, implemented the experiments, analyzed the results, and wrote the
manuscript.

\section*{Conflicts of Interest}
The authors declare no competing interests.



\begin{thebibliography}{99}

\bibitem{vovk2005algorithmic}
Vovk, V.; Gammerman, A.; Shafer, G.
\textit{Algorithmic Learning in a Random World}; Springer: New York, 2005.

\bibitem{angelopoulos2021gentle}
Angelopoulos, A. N.; Bates, S.
A Gentle Introduction to Conformal Prediction and Distribution-Free Uncertainty
Quantification. \textit{Foundations and Trends in Machine Learning} \textbf{2023},
16, 494--591. arXiv:2107.07511.

\bibitem{sadinle2019least}
Sadinle, M.; Lei, J.; Wasserman, L.
Least Ambiguous Set-Valued Classifiers With Bounded Error Levels.
\textit{Journal of the American Statistical Association} \textbf{2019}, 114, 223--234.

\bibitem{romano2020classification}
Romano, Y.; Sesia, M.; Cand\`es, E.
Classification with Valid and Adaptive Coverage.
In \textit{Advances in Neural Information Processing Systems (NeurIPS)}, 2020.

\bibitem{vovk2013conditional}
Vovk, V.
Conditional Validity of Inductive Conformal Predictors.
\textit{Machine Learning} \textbf{2013}, 92, 349--376.

\bibitem{wu2018moleculenet}
Wu, Z.; Ramsundar, B.; Feinberg, E. N.; Gomes, J.; Geniesse, C.; Pappu, A. S.;
Leswing, K.; Pande, V.
MoleculeNet: A Benchmark for Molecular Machine Learning.
\textit{Chemical Science} \textbf{2018}, 9, 513--530.

\bibitem{rogers2010extended}
Rogers, D.; Hahn, M.
Extended-Connectivity Fingerprints.
\textit{Journal of Chemical Information and Modeling} \textbf{2010}, 50, 742--754.

\bibitem{breiman2001random}
Breiman, L.
Random Forests.
\textit{Machine Learning} \textbf{2001}, 45, 5--32.

\bibitem{kipf2017semi}
Kipf, T. N.; Welling, M.
Semi-Supervised Classification with Graph Convolutional Networks.
In \textit{International Conference on Learning Representations (ICLR)}, 2017.

\bibitem{fey2019fast}
Fey, M.; Lenssen, J. E.
Fast Graph Representation Learning with PyTorch Geometric.
\textit{arXiv:1903.02428}, 2019.

\bibitem{chithrananda2020chemberta}
Chithrananda, S.; Grand, G.; Ramsundar, B.
ChemBERTa: Large-Scale Self-Supervised Pretraining for Molecular Property
Prediction. \textit{arXiv:2010.09885}, 2020.

\bibitem{guo2017calibration}
Guo, C.; Pleiss, G.; Sun, Y.; Weinberger, K. Q.
On Calibration of Modern Neural Networks.
In \textit{International Conference on Machine Learning (ICML)}, 2017.

\bibitem{norinder2014introducing}
Norinder, U.; Carlsson, L.; Boyer, S.; Eklund, M.
Introducing Conformal Prediction in Predictive Modeling. A Transparent and Flexible
Alternative to Applicability Domain Determination.
\textit{Journal of Chemical Information and Modeling} \textbf{2014}, 54, 1596--1603.

\bibitem{eklund2015application}
Eklund, M.; Norinder, U.; Boyer, S.; Carlsson, L.
The Application of Conformal Prediction to the Drug Discovery Process.
\textit{Annals of Mathematics and Artificial Intelligence} \textbf{2015}, 74, 117--132.

\bibitem{martins2012bayesian}
Martins, I. F.; Teixeira, A. L.; Pinheiro, L.; Falcao, A. O.
A Bayesian Approach to in Silico Blood--Brain Barrier Penetration Modeling.
\textit{Journal of Chemical Information and Modeling} \textbf{2012}, 52, 1686--1697.

\bibitem{subramanian2016computational}
Subramanian, G.; Ramsundar, B.; Pande, V.; Denny, R. A.
Computational Modeling of $\beta$-Secretase 1 (BACE-1) Inhibitors Using Ligand-Based
Approaches.
\textit{Journal of Chemical Information and Modeling} \textbf{2016}, 56, 1936--1949.

\bibitem{huang2016tox21}
Huang, R.; Xia, M.; Nguyen, D.-T.; Zhao, T.; Sakamuru, S.; Zhao, J.; Shahane, S. A.;
Rossoshek, A.; Simeonov, A.
Tox21Challenge to Build Predictive Models of Nuclear Receptor and Stress Response
Pathways as Mediated by Exposure to Environmental Chemicals and Drugs.
\textit{Frontiers in Environmental Science} \textbf{2016}, 3, 85.

\bibitem{landrum_rdkit}
Landrum, G. RDKit: Open-Source Cheminformatics. \url{https://www.rdkit.org}.

\bibitem{bemis1996properties}
Bemis, G. W.; Murcko, M. A.
The Properties of Known Drugs. 1. Molecular Frameworks.
\textit{Journal of Medicinal Chemistry} \textbf{1996}, 39, 2887--2893.

\end{thebibliography}
\end{document}